\documentclass[10pt,twocolumn,letterpaper]{article}

\usepackage{cvpr}
\usepackage{times}
\usepackage{epsfig}
\usepackage{graphicx}
\usepackage{amsmath}
\usepackage{amssymb}

\usepackage{nicefrac}       
\usepackage{microtype}      

\usepackage{multirow}
\usepackage{verbatim}
\usepackage{color}
\usepackage{float}
\usepackage{enumitem}
\usepackage{booktabs}
\usepackage{tabulary,multirow,overpic,xcolor}
\usepackage[caption=false]{subfig}
\usepackage{pifont}
\usepackage{epstopdf}
\usepackage{t1enc}

\usepackage[british,UKenglish,USenglish,english,american]{babel}

\newcommand{\figref}[1]{Fig.~\ref{#1}}
\newcommand{\tblref}[1]{Table~\ref{#1}}
\newcommand{\sref}[1]{Sec.~\ref{#1}}

\usepackage{soul}
\usepackage{soul}
\definecolor{carmine}{rgb}{0.59, 0.0, 0.09}

\newcommand{\rw}[1]{{#1}}  

\newcommand{\app}{\raise.17ex\hbox{$\scriptstyle\sim$}}

\def\x{$\times$}
\newcolumntype{x}[1]{>{\centering\arraybackslash}p{#1pt}}

\newlength\savewidth\newcommand\shline{\noalign{\global\savewidth\arrayrulewidth
		\global\arrayrulewidth 1pt}\hline\noalign{\global\arrayrulewidth\savewidth}}
\newcommand{\tablestyle}[2]{\setlength{\tabcolsep}{#1}\renewcommand{\arraystretch}{#2}\centering\footnotesize}

\makeatletter\renewcommand\paragraph{\@startsection{paragraph}{4}{\z@}
	{.5em \@plus1ex \@minus.2ex}{-.5em}{\normalfont\normalsize\bfseries}}\makeatother

\definecolor{citecolor}{RGB}{34,139,34}
\usepackage[pagebackref=true,breaklinks=true,letterpaper=true,colorlinks,
citecolor=citecolor,bookmarks=false]{hyperref}

\cvprfinalcopy 

\renewcommand{\omega}{\alpha}
\renewcommand{\phi}{\beta}

\def\cvprPaperID{****} 

\ifcvprfinal\fi  
\begin{document}
	
	
	\title{\vspace{-.5em} SlowFast Networks for Video Recognition \\ \vspace{-.5em}}
	
	\author{
		Christoph Feichtenhofer \qquad
		Haoqi Fan \qquad
		Jitendra Malik \qquad
		Kaiming He \vspace{.8em}\\
		Facebook AI Research (FAIR)
	}
	
	\maketitle
	
	\definecolor{fastcolor}{RGB}{121,178,128}
	\definecolor{slowcolor}{RGB}{165,170,243}
	\newcommand{\fastcolor}[1]{\textcolor{fastcolor}{\textbf{#1}}}
	\newcommand{\fastcolorC}[1]{\textcolor{orange}{#1}}
	\newcommand{\slowcolor}[1]{\textcolor{slowcolor}{#1}}
	
	\newcommand{\slow}{\slowcolor{Slow }}
	\newcommand{\fast}{\fastcolor{Fast }}
	
	\definecolor{predictioncolor}{RGB}{0,255,0}
	\definecolor{labelcolor}{RGB}{255,0,0}
	\newcommand{\predictioncolor}[1]{\textcolor{predictioncolor}{#1}}
	\newcommand{\labelcolor}[1]{\textcolor{labelcolor}{#1}}
	
	\newcommand{\pred}{\predictioncolor{\textbf{Predictions}: }}
	\newcommand{\gt}{\labelcolor{\textbf{Labels}: }}

	\definecolor{demphcolor}{RGB}{144,144,144}
	\newcommand{\demph}[1]{\textcolor{demphcolor}{#1}}
	
	\thispagestyle{empty}

	\begin{abstract}
		\vspace{-.2em}
		We present SlowFast networks for video recognition. Our model involves (i) a Slow pathway, operating at low frame rate, to capture spatial semantics, and (ii) a Fast pathway, operating at high frame rate, to capture motion at fine temporal resolution. The Fast pathway can be made very lightweight by reducing its channel capacity, yet can learn useful temporal information for video recognition.
		Our models achieve strong performance for both action classification and detection in video, and large improvements are pin-pointed as contributions by our SlowFast concept. 
		We report state-of-the-art accuracy on major video recognition benchmarks, Kinetics, Charades and AVA. Code has been made available at: \url{https://github.com/facebookresearch/SlowFast}. 
	\end{abstract}
	
	\section{Introduction}
	\label{sec:introduction}
	
	It is customary in the recognition of images $I(x,y)$ to treat the two spatial dimensions $x$ and $y$ symmetrically. This is justified by the statistics of natural images, which are to a first approximation isotropic---all orientations are equally likely---and shift-invariant \cite{Ruderman1994,Huang1999}. But what about video signals $I(x,y,t)$?  Motion is the spatiotemporal counterpart of orientation \cite{Adelson1985}, but all spatiotemporal orientations are {\em not} equally likely. Slow motions are more likely than fast motions (indeed most of the world we see is at rest at a given moment) and this has been exploited in Bayesian accounts of how humans perceive motion stimuli \cite{Weiss2002}. For example, if we see a moving edge in isolation, we perceive it as moving perpendicular to itself, even though in principle it could also have an arbitrary component of movement tangential to itself (the aperture problem in optical flow). This percept is rational if the prior  favors slow movements. 
	
	If all spatiotemporal orientations are not equally likely, then there is no reason for us to treat space and time symmetrically, as is implicit in approaches to video recognition based on spatiotemporal convolutions \cite{Tran2015,Carreira2017}. We might instead ``factor'' the architecture to treat spatial structures and temporal events separately. For concreteness, let us study this in the context of recognition.
	The categorical spatial semantics of the visual content often evolve \emph{slowly}. For example, waving hands do not change their identity as ``hands'' over the span of the waving action, and a person is always in the ``person'' category even though he/she can transit from walking to running. So the recognition of the categorical semantics (as well as their colors, textures, lighting \etc) can be refreshed relatively \emph{slowly}. On the other hand, the motion being performed can evolve much \emph{faster} than their subject identities, such as clapping, waving, shaking, walking, or jumping. It can be desired to use \emph{fast} refreshing frames (high temporal resolution) to effectively model the potentially \emph{fast} changing motion.

	\begin{figure}[t]
		\centering
		\vspace{2em}
		\includegraphics[width=1\linewidth]{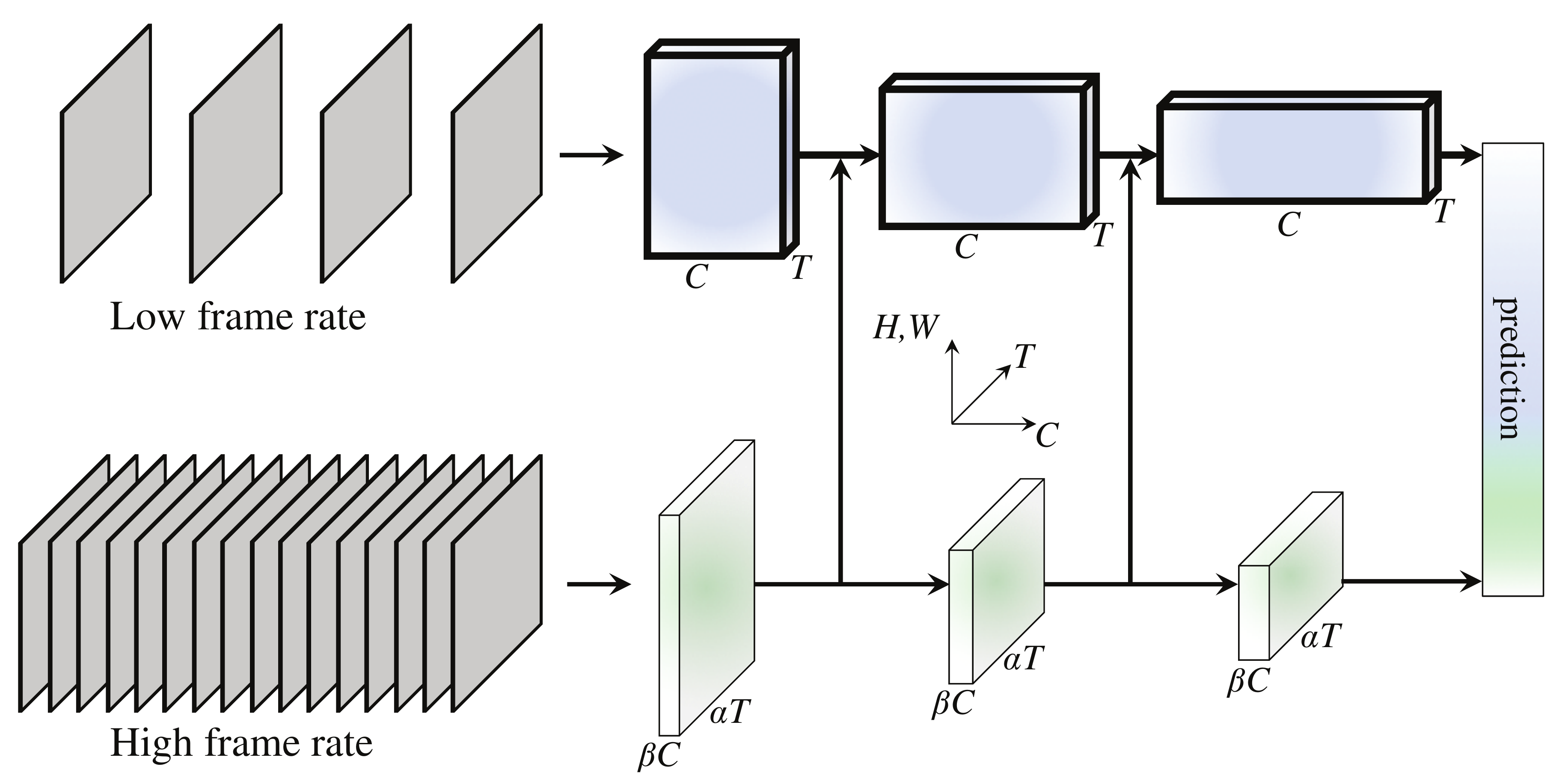}
		\caption{\textbf{A SlowFast network} has a low frame rate, low temporal resolution \emph{Slow} pathway and a high frame rate, $\omega$$\times$ higher temporal resolution \emph{Fast} pathway. The Fast pathway is lightweight by using a  fraction ($\phi$, \eg, 1/8) of channels. Lateral connections fuse them. 
		}
		\label{fig:teaser}
	\end{figure}
	
	Based on this intuition, we present a two-pathway \emph{SlowFast} model for video recognition (\figref{fig:teaser}). 
	One pathway is designed to capture semantic information that can be given by images or a few sparse frames, and it operates at \emph{low} frame rates and \emph{slow} refreshing speed.
	In contrast, the other pathway is responsible for capturing rapidly changing motion, by operating at \emph{fast} refreshing speed and high temporal resolution.
	Despite its high temporal rate, this pathway is made very \emph{lightweight}, \eg, $\app$20\% of total computation.
	This is because this pathway is designed to have fewer channels and weaker ability to process spatial information, while such information can be provided by the first pathway in a less redundant manner.
	We call the first a \emph{Slow} pathway and the second a \emph{Fast} pathway, driven by their different temporal speeds.
	The two pathways are fused by lateral connections.
	
	Our conceptual idea leads to flexible and effective designs for video models. The Fast pathway, due to its lightweight nature, does \emph{not} need to perform any temporal pooling---it can operate on high frame rates for all intermediate layers and maintain temporal fidelity. Meanwhile, thanks to the lower temporal rate, the Slow pathway can be more focused on the spatial domain and semantics. By treating the raw video at different temporal rates, our method allows the two pathways to have their own expertise on video modeling.

	There is another well known architecture for video recognition which has a two-stream design \cite{Simonyan2014}, but provides conceptually different perspectives. The Two-Stream method \cite{Simonyan2014} has not explored the potential of \emph{different temporal speeds}, a key concept in our method.
	The two-stream method adopts the same backbone structure to both streams, whereas our Fast pathway is more lightweight.
	Our method does not compute optical flow, and therefore, our models are learned end-to-end from the raw data. In our experiments we observe that the SlowFast network is empirically more effective.

	Our method is partially inspired by biological studies on the retinal ganglion cells in the primate visual system \cite{Hubel1965,Livingstone1988,Derrington1984,Felleman1991,VanEssen1994}, though admittedly the analogy is rough and premature. These studies found that in these cells, $\app$80\% are Parvocellular (P-cells) and $\app$15-20\% are Magnocellular (M-cells).
	The M-cells operate at \emph{high temporal frequency} and are responsive to fast temporal changes, but not sensitive to spatial detail or color.
	P-cells provide fine spatial detail and color, but lower temporal resolution, responding slowly to stimuli.
	Our framework is analogous in that: (i) our model has two pathways separately working at low and high temporal resolutions; (ii) our Fast pathway is designed to capture fast changing motion but fewer spatial details, analogous to M-cells; and (iii) our Fast pathway is lightweight, similar to the small ratio of M-cells. We hope these relations will inspire more  computer vision models for video recognition.
	
	We  evaluate our method on the Kinetics-400 \cite{Kay2017}, Kinetics-600 \cite{Carreira2018}, Charades \cite{Sigurdsson2016} and AVA \cite{Gu2018} datasets. Our comprehensive ablation experiments 	on Kinetics action classification demonstrate the efficacy contributed by SlowFast.  SlowFast networks set a new state-of-the-art on all datasets with significant gains to previous systems in the literature.
	
	\section{Related Work}
	\label{sec:related_work}
	
	\paragraph{Spatiotemporal filtering.}
	Actions can be formulated as spatiotemporal objects and captured by oriented filtering in spacetime, as done by HOG3D \cite{Klaeser2008} and cuboids \cite{Dollar2005}.
	3D ConvNets \cite{Taylor2010,Tran2015,Carreira2017} extend 2D image models \cite{Krizhevsky2012,Simonyan2015,Szegedy2015,He2016} to the spatiotemporal domain, handling both spatial and temporal dimensions similarly.
	There are also related methods focusing on long-term filtering and pooling using temporal strides \cite{Varol2018,Feichtenhofer2016,Wang2016a,Zhou2017}, as well as decomposing the convolutions into separate 2D spatial and 1D temporal filters \cite{Feichtenhofer2016a,Tran2018,Xie2018,Qiu2017}.
	
	Beyond spatiotemporal filtering or their separable versions, our work pursuits a more thorough separation of modeling expertise by using two different temporal speeds.
	
	\paragraph{Optical flow for video recognition.}
	There is a classical branch of research focusing on hand-crafted spatiotemporal features based on optical flow.
	These methods, including histograms of flow \cite{Laptev2008}, motion boundary histograms \cite{Dalal2006}, and trajectories \cite{Wang2013}, had shown competitive performance for action recognition before the prevalence of deep learning.
	
	In the context of deep neural networks, the two-stream method \cite{Simonyan2014} exploits optical flow by viewing it as another input modality. This method has been a foundation of many competitive results in the literature \cite{Feichtenhofer2016a,Feichtenhofer2016,Wang2016a}.
	However, it is methodologically unsatisfactory given that optical flow is a hand-designed representation, and two-stream methods are often not learned end-to-end jointly with the flow.  
	
	\section{SlowFast Networks}  \label{sec:slow_fast}
	\rw{SlowFast networks can be described as a single stream architecture that operates at two different framerates, but we use the concept of pathways to reflect analogy with the biological Parvo- and Magnocellular counterparts.}
	Our generic architecture has a Slow pathway (\sref{sec:slow}) and a Fast pathway (\sref{sec:fast}), which are fused by lateral connections to a SlowFast network (\sref{sec:lateral}). \figref{fig:teaser} illustrates our concept.
	
	\subsection{Slow pathway} \label{sec:slow}
	
	The Slow pathway can be any convolutional model (\eg, \cite{Feichtenhofer2016a,Tran2015,Carreira2017,Wang2018}) that works on a clip of video as a spatiotemporal volume.
	The key concept in our Slow pathway is a \emph{large} temporal stride $\tau$ on input frames, \ie, it processes only one out of $\tau$ frames.  
	A typical value of $\tau$ we studied is 16---this refreshing speed is roughly 2 frames sampled per second for 30-fps videos.
	Denoting the number of frames sampled by the Slow pathway as $T$, the raw clip length is $T\times\tau$ frames.
	
	\subsection{Fast pathway} \label{sec:fast}
	In parallel to the Slow pathway, the Fast pathway is another convolutional model with the following properties.
	
	\paragraph{High frame rate.}
	Our goal here is to have a fine representation along the temporal dimension. 
	Our Fast pathway works with a \emph{small} temporal stride of $\tau / \omega$, where $\omega > 1$ is the frame rate ratio between the Fast and Slow pathways.
	The two pathways operate on the same raw clip, so the Fast pathway samples $\omega T$ frames, $\omega$ times denser than the Slow pathway. A typical value is $\omega=8$ in our experiments.
	
	The presence of $\omega$ is in the key of the SlowFast concept (\figref{fig:teaser}, time axis). It explicitly indicates that the two pathways work on \emph{different} temporal speeds, and thus drives the expertise of the two subnets instantiating the two pathways. 
	
	\paragraph{High temporal resolution features.} Our Fast pathway not only has a high input resolution, but also pursues high-resolution features throughout the network hierarchy. In our instantiations, we use \emph{no} temporal downsampling layers (neither temporal pooling nor time-strided convolutions) throughout the Fast pathway, until the global pooling layer before classification. As such, our feature tensors always have $\omega T$ frames along the temporal dimension, maintaining temporal fidelity as much as possible.
	
	\paragraph{Low channel capacity.} Our Fast pathway also distinguishes with existing models in that it can use significantly \emph{lower} channel capacity to achieve good accuracy for the SlowFast model. This makes it lightweight.
	
	In a nutshell, our Fast pathway is a convolutional network analogous to the Slow pathway, but has a ratio of $\phi$ ($\phi<1$) channels of the Slow pathway. The typical value is $\phi=1/8$ in our experiments. Notice that the computation (floating-number operations, or FLOPs) of a common layer is often \emph{quadratic} in term of its channel scaling ratio. This is what makes the Fast pathway more computation-effective than the Slow pathway. In our instantiations, the Fast pathway typically takes $\app$20\% of the total computation. Interestingly, as mentioned in \sref{sec:introduction}, evidence suggests that $\app$15-20\% of the retinal cells in the primate visual system are M-cells (that are sensitive to fast motion but not color or spatial detail).
	
	The low channel capacity can also be interpreted as a \emph{weaker} ability of representing spatial semantics. Technically, our Fast pathway has no special treatment on the spatial dimension, so its spatial modeling capacity should be lower than the Slow pathway because of fewer channels. The good results of our model suggest that it is a desired tradeoff for the Fast pathway to weaken its spatial modeling ability while strengthening its temporal modeling ability.
	
	Motivated by this interpretation, we also explore different ways of weakening spatial capacity in the Fast pathway, including reducing input spatial resolution and removing color information. As we will show by experiments, these versions can all give good accuracy, suggesting that a lightweight Fast pathway with less spatial capacity can be made beneficial.
	
	\subsection{Lateral connections} \label{sec:lateral}
	
	The information of the two pathways is fused, so one pathway is not unaware of the representation learned by the other pathway.
	We implement this by \emph{lateral connections}, which have been used to fuse optical flow-based, two-stream networks 
	\cite{Feichtenhofer2016a,Feichtenhofer2016}.
	In image object detection, lateral connections \cite{Lin2017} are a popular technique for merging different levels of spatial resolution and semantics. 
	
	Similar to \cite{Feichtenhofer2016a,Lin2017}, we attach one lateral connection between the two pathways for every ``stage" (\figref{fig:teaser}). Specifically for ResNets \cite{He2016}, these connections are right after pool$_1$, res$_2$, res$_3$, and res$_4$. The two pathways have different temporal dimensions, so the lateral connections perform a transformation to match them (detailed in \sref{sec:realization}).
	We use unidirectional connections that fuse features of the Fast pathway into the Slow one (\figref{fig:teaser}). We have experimented with bidirectional fusion and found similar results.
	
	Finally, a global average pooling is performed on each pathway's output. Then two pooled feature vectors are concatenated as the input to the fully-connected classifier layer.
	
	\newcommand{\blocks}[3]{\multirow{3}{*}{\(\left[\begin{array}{c}\text{1$\times$1$^\text{2}$, #2}\\[-.1em] \text{1$\times$3$^\text{2}$, #2}\\[-.1em] \text{1$\times$1$^\text{2}$, #1}\end{array}\right]\)$\times$#3}
	}
	\newcommand{\blockt}[3]{\multirow{3}{*}{\(\left[\begin{array}{c}\text{\underline{3$\times$1$^\text{2}$}, #2}\\[-.1em] \text{1$\times$3$^\text{2}$, #2}\\[-.1em] \text{1$\times$1$^\text{2}$, #1}\end{array}\right]\)$\times$#3}
	}
	\newcommand{\outsizes}[7]{\multirow{#7}{*}{\(\begin{array}{c} \text{\emph{Slow}}: \text{#1$\times$#2$^\text{2}$}\\[-.1em] \text{\emph{Fast}}: \text{#4$\times$#5$^\text{2}$}\end{array}\)}
	}
	
	\begin{table}[t]
			\vspace{-.3em}
		\scriptsize
		\centering
		\resizebox{\columnwidth}{!}{
			\tablestyle{1pt}{1.08}
			\begin{tabular}{c|c|c|c}
				stage & \emph{Slow} pathway &  \emph{Fast} pathway & output sizes $T$\x$S^2$ \\
				\shline
				\multirow{1}{*}{raw clip} & - & - & 64\x224$^\text{2}$ \\
				\hline
				\multirow{2}{*}{data layer} & \multirow{2}{*}{stride 16, 1$^\text{2}$} & \multirow{2}{*}{stride \textbf{2}, 1$^\text{2}$} &  \outsizes{{4}}{224}{64}{\fastcolor{32}}{224}{8}{2}   \\
				&  &  \\
				\hline
				\multirow{2}{*}{conv$_1$} & \multicolumn{1}{c|}{1\x7$^\text{2}$, {64}} & \multicolumn{1}{c|}{\underline{5\x7$^\text{2}$}, \fastcolorC{8}} &  \outsizes{{4}}{112}{64}{\fastcolor{32}}{112}{8}{2}   \\
				& stride 1, 2$^\text{2}$ & stride 1, 2$^\text{2}$  \\
				\hline
				\multirow{2}{*}{pool$_1$}  & \multicolumn{1}{c|}{1\x3$^\text{2}$ max} & \multicolumn{1}{c|}{1\x3$^\text{2}$ max} &  \outsizes{{4}}{56}{64}{\fastcolor{32}}{56}{8}{2} \\
				& stride 1, 2$^\text{2}$ & stride 1, 2$^\text{2}$ & \\
				\hline
				\multirow{3}{*}{res$_2$} & \blocks{{256}}{{64}}{3} & \blockt{\fastcolorC{32}}{\fastcolorC{8}}{3} & \outsizes{{4}}{56}{256}{\fastcolor{32}}{56}{32}{3}  \\
				&  & \\
				&  & \\
				\hline
				\multirow{3}{*}{res$_3$} & \blocks{{512}}{{128}}{4} &  \blockt{\fastcolorC{64}}{\fastcolorC{16}}{4}  & \outsizes{{4}}{28}{512}{\fastcolor{32}}{28}{64}{3}  \\
				&  & \\
				&  & \\
				\hline
				\multirow{3}{*}{res$_4$} & \blockt{{1024}}{{256}}{6} & \blockt{\fastcolorC{128}}{\fastcolorC{32}}{6} &  \outsizes{{4}}{14}{1024}{\fastcolor{32}}{14}{128}{3}  \\
				&  & \\
				&  & \\
				\hline
				\multirow{3}{*}{res$_5$} & \blockt{{2048}}{{512}}{3} & \blockt{\fastcolorC{256}}{\fastcolorC{64}}{3} &   \outsizes{{4}}{7}{2048}{\fastcolor{32}}{7}{256}{3} \\
				&  & \\
				&  & \\
				\hline
				\multicolumn{3}{c|}{global average pool, concate, fc}  & \# classes \\
		\end{tabular}}
		\vspace{.1em}
		\caption{\textbf{An example instantiation of the SlowFast network}. 
			The dimensions of kernels are denoted by $\{$$T$\x $S^2$, $C$$\}$ for temporal, spatial, and channel sizes.
			Strides are denoted as $\{$temporal stride, spatial stride$^2$$\}$.
			Here the speed ratio is $\omega=8$ and the channel ratio is $\phi=1/8$. $\tau$ is 16. The \fastcolor{green} colors mark \emph{higher} temporal resolution, and \fastcolorC{orange} colors mark \emph{fewer} channels, for the Fast pathway. 
			Non-degenerate temporal filters are underlined.
			Residual blocks are shown by brackets. The backbone is ResNet-50.
		}
		\label{tab:arch}
			\vspace{-.8em}
	\end{table}
	
	\subsection{Instantiations}  \label{sec:realization}
	
	Our idea of SlowFast is generic, and it can be instantiated with different backbones (\eg, \cite{Simonyan2015,Szegedy2015,He2016}) and implementation specifics. 
	In this subsection, we describe our instantiations of the network architectures.
	
	An example SlowFast model is specified in Table~\ref{tab:arch}.
	We denote spatiotemporal size by $T$\x $S^2$ where $T$ is the temporal length and $S$ is the height and width of a square spatial crop. The details are described next.
	
	\paragraph{Slow pathway.}
	The Slow pathway in Table~\ref{tab:arch} is a temporally strided 3D ResNet, modified from \cite{Feichtenhofer2016a}. It has $T$ $=$ 4 frames as the network input, sparsely sampled from a 64-frame raw clip with a temporal stride $\tau$ $=$ 16. We opt to not perform temporal downsampling in this instantiation, as doing so would be detrimental when the input stride is large.
	
	Unlike typical C3D / I3D models, we use \emph{non-degenerate} temporal convolutions (temporal kernel size $>$ 1, underlined in Table~\ref{tab:arch}) only in res$_4$ and res$_5$; all filters from conv$_1$ to res$_3$ are essentially 2D convolution kernels in this pathway. This is motivated by our experimental observation that using temporal convolutions in earlier layers degrades accuracy. We argue that this is because when objects move fast and the temporal stride is large, there is little correlation within a temporal receptive field unless the spatial receptive field is large enough (\ie, in later layers).
	
	\paragraph{Fast pathway.}
	Table~\ref{tab:arch} shows an example of the Fast pathway with $\omega=8$ and $\phi=1/8$. It has a much higher temporal resolution (\fastcolor{\textbf{green}}) and lower channel capacity (\fastcolorC{orange}).
	
	The Fast pathway has non-degenerate temporal convolutions in \emph{every} block. This is motivated by the observation that this pathway holds fine temporal resolution for the temporal convolutions to capture detailed motion. Further, the Fast pathway has no temporal downsampling layers by design.
	
	\paragraph{Lateral connections.} Our lateral connections fuse from the Fast to the Slow pathway.
	It requires to match the sizes of features before fusing.
	Denoting the feature shape of the Slow pathway as $\{$$T$, $S^2$, $C$$\}$, the feature shape of the Fast pathway is $\{$$\omega T$, $S^2$, $\phi C$$\}$.
	We experiment with the following transformations in the lateral connections:
	
	\vspace{.5em}
	\noindent{(i) \emph{Time-to-channel}:} We reshape and transpose $\{$$\omega T$, $S^2$, $\phi C$$\}$ into $\{$$T$, $S^2$, $\omega\phi C$$\}$, meaning that we pack all $\omega$ frames into the channels of one frame.
	
	\vspace{.5em}
	\noindent{(ii) \emph{Time-strided sampling}:} We simply sample one out of every $\omega$ frames, so $\{$$\omega T$, $S^2$, $\phi C$$\}$ becomes $\{$$T$, $S^2$, $\phi C$$\}$.
	
	\vspace{.5em}
	\noindent{(iii) \emph{Time-strided convolution}:} We perform a 3D convolution of a 5\x1$^\text{2}$ kernel with $2\phi C$ output channels and stride $=$ $\omega$.
	
	\vspace{.5em}
	\noindent
	The output of the lateral connections is fused into the Slow pathway by summation or concatenation.
	
	\section{Experiments: Action Classification} \label{sec:kinetics}
	
	We evaluate our approach on four video recognition datasets using standard evaluation protocols. For the action classification experiments, presented in this section we consider the widely used Kinetics-400  \cite{Kay2017}, the recent Kinetics-600 \cite{Carreira2018}, and Charades \cite{Sigurdsson2016}. For action detection experiments in \sref{sec:detection}, we use the challenging AVA dataset \cite{Gu2018}. 
	
		\paragraph{Training.} Our models on Kinetics are trained \emph{from random initialization} (``\emph{from scratch}''), \emph{without} using ImageNet \cite{Deng2009} or any pre-training. We use synchronized SGD training following the recipe in \cite{Goyal2017}. See details in Appendix.
	
	For the temporal domain, we randomly sample a clip (of $\alpha T$\x$\tau$ frames) from the full-length video, and the input to the Slow and Fast pathways are respectively $T$ and $\omega T$ frames; for the spatial domain, we randomly crop 224\x224 pixels from a video, or its horizontal flip, with a shorter side randomly sampled in [256, 320] pixels \cite{Simonyan2015,Wang2018}. 
 
	\paragraph{Inference.}  Following common practice, we uniformly sample 10 clips from a video along its temporal axis.
	For each clip, we scale the shorter spatial side to 256 pixels and take 3 crops of 256\x256 to cover the spatial dimensions, as an approximation of fully-convolutional testing, following the code of \cite{Wang2018}. We average the softmax scores for prediction.
	
	We report the actual \emph{inference-time} computation. As {existing papers differ in their inference strategy for cropping/clipping in space and in time}. When comparing to previous work, we report the FLOPs per spacetime ``view" (temporal clip with spatial crop) at inference \emph{and} the number of views used. Recall that in our case, the inference-time spatial size is 256$^2$ (instead of 224$^2$ for training) and 10 temporal clips each with 3 spatial crops are used (30 views).

	\paragraph{Datasets.} Kinetics-400 \cite{Kay2017} consists of $\app$240k training videos and 20k validation videos in 400 human action categories.
Kinetics-600 \cite{Carreira2018} has $\app$392k training videos and 30k validation videos in 600 classes. 
We report top-1 and top-5 classification accuracy (\%). We report the computational cost (in FLOPs) of a single, spatially center-cropped clip.
	
	Charades \cite{Sigurdsson2016} has $\app$9.8k training videos and 1.8k validation videos in 157 classes in a multi-label classification setting of longer activities spanning $\app$30 seconds on average. Performance is measured in mean Average Precision (mAP).
	
	\subsection{Main Results}
		
\paragraph{Kinetics-400.} \tblref{tab:sota:k400} shows the comparison with state-of-the-art results for our SlowFast instantiations using various input samplings ($T$\x$\tau$) and backbones: ResNet-50/101 (R50/101) \cite{He2016} and Nonlocal (NL) \cite{Wang2018}.

	In comparison to the previous state-of-the-art \cite{Wang2018} our best  model provides 2.1\% higher top-1 accuracy. 
	Notably, all our results are substantially better than existing results that are also \emph{without ImageNet pre-training}. In particular, our model (79.8\%) is \textbf{5.9\%} absolutely better than the previous best result of this kind (73.9\%). We have experimented with ImageNet pretraining for SlowFast networks and found that they perform similar ($\pm$0.3\%) for both the pre-trained and the train from scratch (random initialization) variants.
	
	\begin{table}[t!]
				\vspace{-1.5em}
		\centering
		\small
		\tablestyle{1.8pt}{1.05}
		\begin{tabular}{l|c|c|c|c|c}
			\multicolumn{1}{c|}{model} & flow  &\multicolumn{1}{c|}{pretrain} &   top-1  & top-5  & GFLOPs\x views  \\
			\shline
			\demph{I3D \cite{Carreira2017}} && \demph{ImageNet} & \demph{72.1} &  \demph{90.3} & \demph{108 \x} \demph{N/A}  \\
			\demph{Two-Stream I3D \cite{Carreira2017}}&  \checkmark  &  \demph{ImageNet} &   \demph{75.7} &  \demph{92.0} & \demph{216~\x} \demph{N/A}  \\
			\demph{S3D-G \cite{Xie2018}}  & \checkmark &  \demph{ImageNet} &   \demph{77.2} & \demph{93.0} &  \demph{143~\x}  \demph{N/A}  \\
			\demph{Nonlocal R50 \cite{Wang2018}}  &  &  \demph{ImageNet} &   \demph{76.5} & \demph{92.6} &  \demph{282~\x}  \demph{30}  \\
			\demph{Nonlocal R101 \cite{Wang2018}} &  &  \demph{ImageNet} &   \demph{77.7} & \demph{93.3} & \demph{359~\x}  \demph{30}  \\
			\hline
			R(2+1)D Flow \cite{Tran2018} & \checkmark & -&   67.5 &  87.2 & 152~\x~115  \\
			STC \cite{Diba2018} &  &  -&  68.7 &  88.5  & N/A~\x~N/A  \\
			ARTNet \cite{Wang2018a} &  &  -&    69.2 & 88.3 & 23.5~\x~250 \\
			{S3D \cite{Xie2018}}  &  & -&   {69.4} & {89.1} &  {66.4~\x}  {N/A}  \\
			ECO \cite{Zolfaghari2018}&   &  - &  70.0 &  89.4 & N/A~\x~N/A  \\
			
			I3D \cite{Carreira2017}&  \checkmark &  -&    71.6 & 90.0 & 216~\x~N/A  \\
			
			R(2+1)D \cite{Tran2018}&   &  -&   72.0 &  90.0 & 152~\x~115  \\
			R(2+1)D \cite{Tran2018} & \checkmark &  -&  73.9 &  90.9& 304~\x~115  \\
			\hline
			\textbf{SlowFast} 4\x 16, R50&  &- & 75.6  & 92.1 & 36.1~\x~30 \\
			\textbf{SlowFast} 8\x 8, R50 &  &- & 77.0  & 92.6 & 65.7~\x~30  \\
			\textbf{SlowFast} 8\x 8, R101 &   &- & 77.9  & 93.2 & 106~\x~30  \\
			\textbf{SlowFast} 16\x 8, R101   &  &- & {78.9} & {93.5} &  213~\x~30  \\
			\textbf{SlowFast} 16\x 8, R101+NL  &  &- &\textbf{79.8} & \textbf{93.9} &  234~\x~30  \\
		\end{tabular}
		\vspace{.1em}
		\caption{\textbf{Comparison with the state-of-the-art on Kinetics-400}. In the last column, we report the inference cost with a single ``view" (temporal clip with spatial crop) $\times$ the numbers of such views used. The SlowFast models are with different input sampling ($T$\x$\tau$) and backbones (R-50, R-101, NL). ``N/A'' indicates the numbers are not available for us.  
		}
		\label{tab:sota:k400}
		\vspace{-.5em}
	\end{table}
	
	\begin{figure}[t]
			\vspace{-0.5em}
		\centering
		\includegraphics[width=1\linewidth]{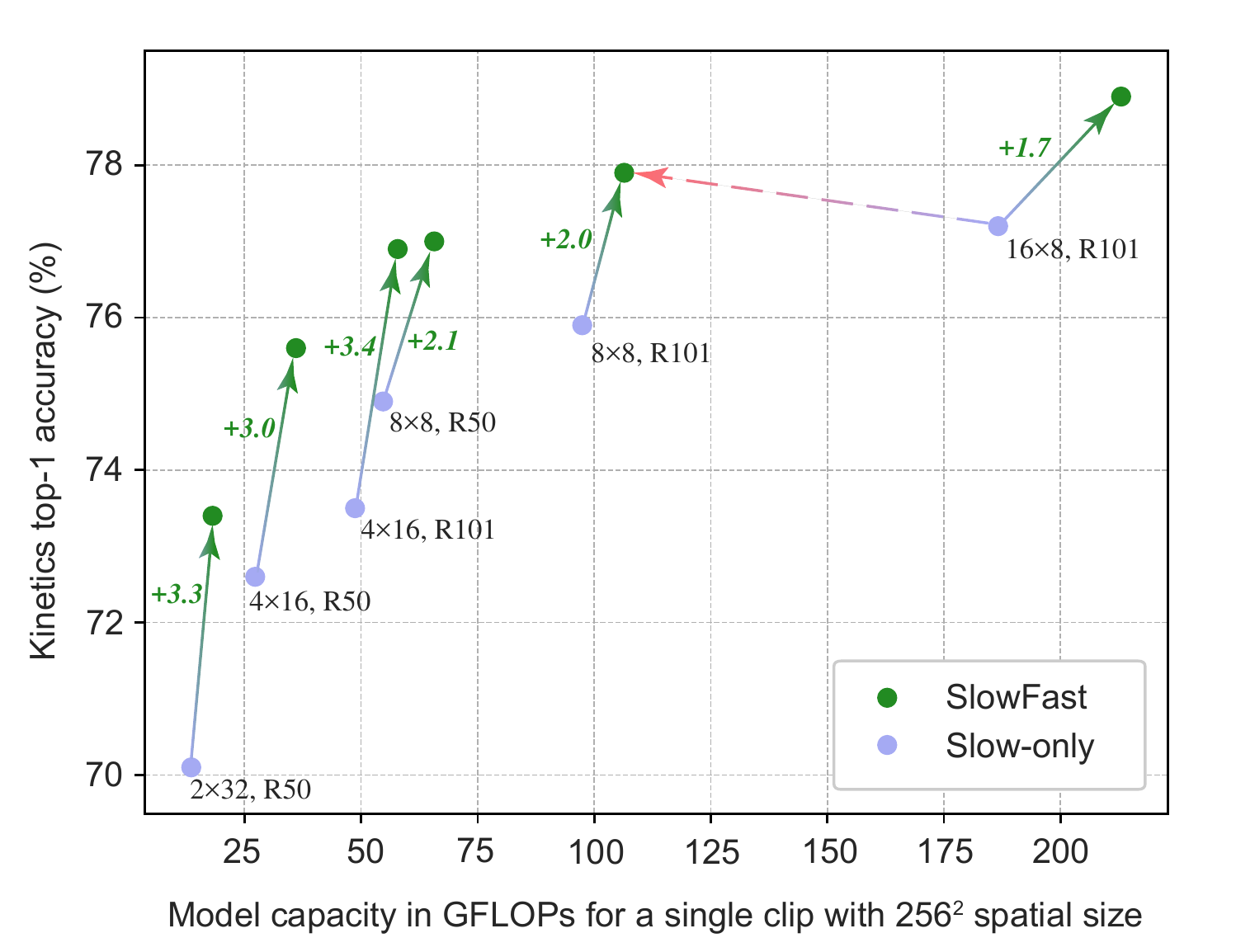}
		\vspace{-1.3em}
		\caption{\textbf{Accuracy/complexity tradeoff} on Kinetics-400 for the SlowFast (green) \vs  Slow-only (blue) architectures. 
		SlowFast is consistently better than its Slow-only counterpart in all cases (green arrows). SlowFast provides higher accuracy \emph{and} lower cost than temporally heavy Slow-only (\eg red arrow).
		The complexity is for a single 256$^2$ view, and accuracy are obtained by 30-view testing.  
		}
		\label{fig:slowVsSlowFast}
		\vspace{-0.3em}
	\end{figure}
	

		Our results are achieved at low inference-time cost. We notice that many existing works (if reported) use \emph{extremely dense} sampling of clips along the temporal axis, which can lead to $>$100 views at inference time. This cost has been largely overlooked. In contrast, our method does not require many temporal clips, due to the high temporal resolution yet lightweight Fast pathway. Our cost per spacetime view can be low (\eg, 36.1 GFLOPs), while still being accurate. 
	
	The SlowFast variants from \tblref{tab:sota:k400} (with different backbones and sample rates) are compared in \figref{fig:slowVsSlowFast} the  with their corresponding Slow-only pathway to assess the improvement brought by the Fast pathway. The horizontal axis measures model capacity  for a single input clip of 256$^2$ spatial size, which is proportional to 1$/$30 of the overall inference cost. \figref{fig:slowVsSlowFast} shows that for all variants the Fast pathway is able to consistently improve the performance of the Slow counterpart at comparatively low cost. The next subsection provides a more detailed analysis on Kinetics-400.
	
	\paragraph{Kinetics-600} is relatively new, and existing results are limited. So our goal is mainly to provide results for future reference in  \tblref{tab:sota:k600}.  Note that the Kinetics-600 validation set overlaps with the Kinetics-400 training set \cite{Carreira2018}, and therefore we do \emph{not} pre-train on Kinetics-400. 
	The winning entry \cite{He2018} of the latest ActivityNet Challenge 2018 \cite{Ghanem2018} reports a best single-model, single-modality accuracy of 79.0\%. Our variants show good performance with the best model at 81.8\%. SlowFast results on the  recent Kinetics-700 \cite{Carreira2019} are in \cite{Feichtenhofer2019a}. 
	
	\begin{table}[t]
		\vspace{-1.5em}
		\centering
		\small
		\tablestyle{2pt}{1.05}
		\begin{tabular}{l|c|c|c|c}
			\multicolumn{1}{c|}{model} &\multicolumn{1}{c|}{pretrain} &   top-1  & top-5    & \scriptsize GFLOPs\x views    \\ 
			\shline
			
			I3D \cite{Carreira2018} & - &   71.9 & 90.1  &  108~\x~N/A  \\
			\demph{StNet-IRv2 RGB \cite{He2018}} & \demph{\scriptsize ImgNet+Kin400} &    \demph{79.0} & \demph{N/A} &  \demph{N/A}  \\
			\hline
			\textbf{SlowFast} 4\x 16, R50&  - & 78.8  & 94.0 & 36.1~\x~30 \\
			\textbf{SlowFast} 8\x8, R50 & - & 79.9  & 94.5 &  65.7~\x 30 \\
			\textbf{SlowFast} 8\x8, R101 & - & {80.4}  & {94.8} &  106~\x~30  \\
			\textbf{SlowFast} 16\x8, R101 & - & {81.1}  & {95.1} &  213~\x~30  \\
			\textbf{SlowFast} 16\x8, R101+NL & - & \textbf{81.8}  & \textbf{95.1} &  234~\x~30  \\
		\end{tabular}
		\vspace{.1em}
		\caption{\textbf{Comparison with the state-of-the-art on Kinetics-600}.   SlowFast models the same as in \tblref{tab:sota:k400}.
		}
		\label{tab:sota:k600}
	\end{table}
	
	\begin{table}[t!] 
		\vspace{-.5em}
		\centering
		\small
		\tablestyle{2pt}{1.05}
		\begin{tabular}{l|x{64}|x{22}|c}
			\multicolumn{1}{c|}{model} &  \multicolumn{1}{c|}{pretrain} &  mAP    & \scriptsize GFLOPs\x views    \\ 
			\shline
			CoViAR, R-50 \cite{Wu2018}  &  {\scriptsize ImageNet} &  21.9 & N/A \\
			Asyn-TF, VGG16 \cite{Sigurdsson2017} &  {\scriptsize ImageNet}  &  22.4 & N/A \\
			MultiScale TRN~\cite{Zhou2017} &  {\scriptsize ImageNet}  &  25.2 & N/A \\
			Nonlocal, R101 \cite{Wang2018} &  {\scriptsize ImageNet+Kinetics400}  &   37.5 & 544~\x~30  \\
			STRG, R101+NL \cite{Wang2018b} &  {\scriptsize ImageNet+Kinetics400} &   39.7 & 630~\x~30  \\
			\hline
			our baseline (Slow-only) & {\scriptsize Kinetics-400} & 39.0 &  187~\x~30  \\
			\textbf{SlowFast}  & {\scriptsize Kinetics-400} & 42.1 &  213~\x~30  \\
			\textbf{SlowFast}, +NL & {\scriptsize Kinetics-400} & 42.5  &  234~\x~30  \\
			\textbf{SlowFast}, +NL &  {\scriptsize Kinetics-600} &  \textbf{45.2} &  234~\x~30  \\
		\end{tabular}
		\vspace{.5em}
		\caption{\textbf{Comparison with the state-of-the-art on Charades}. All our variants are based on  $T$\x$\tau$  $=$ 16\x8, R-101.} 	\vspace{-1em}
		\label{tab:sota:charades}
	\end{table} 	

	\begin{table*}[t]\centering\vspace{-1.5em}
		\captionsetup[subfloat]{captionskip=2pt}
		\captionsetup[subffloat]{justification=centering}
		\subfloat[\textbf{SlowFast fusion}: Fusing Slow and Fast pathways with various types of lateral connections throughout the network hierarchy is consistently better than the Slow and Fast only baselines. 
		\label{tab:ablation:fusion}]{
			\tablestyle{2pt}{1.05}
			\begin{tabular}{c|l|x{22}x{22}x{28}}
				&  \multicolumn{1}{c|}{lateral}  & top-1 & top-5 & GFLOPs \\
				\shline
				Slow-only & \multicolumn{1}{c|}{-} & 72.6& 90.3 & 27.3 \\ 
				Fast-only & \multicolumn{1}{c|}{-} & 51.7 & 78.5 &\textbf{ 6.4} \\ 
				\hline
				SlowFast & \multicolumn{1}{c|}{-}  & 73.5 & 90.3 & 34.2 \\ 
				SlowFast & TtoC, sum &  74.5 & 91.3 & 34.2 \\ 
				SlowFast & TtoC, concat & 74.3 & 91.0 & 39.8 \\ 
				SlowFast &T-sample  & 75.4 &  91.8 & 34.9 \\ 
				SlowFast & T-conv& \textbf{75.6} & \textbf{92.1} & 36.1 \\ 
		\end{tabular}}\hspace{3mm}
		\subfloat[\textbf{Channel capacity ratio}: Varying values of $\phi$, the channel capacity ratio of the Fast pathway to make SlowFast lightweight.   \label{tab:ablation:widthFast}]{
			\tablestyle{2pt}{1.05}
			\begin{tabular}{r|x{20}x{20}x{28}}
				\multicolumn{1}{c|}{}  & top-1 & top-5 & GFLOPs \\
				\shline
				Slow-only & 72.6& 90.3 & 27.3 \\ 
				\hline
				$\phi$ $=$ $1/4$ &  75.6 &   91.7 & 54.5 \\ 
				$1/6$ &  \textbf{75.8}  &  92.0 & 41.8 \\ 
				$1/8$ &  75.6 &  \textbf{92.1} & 36.1 \\ 
				$1/12$ & 75.2   & 91.8 & 32.8 \\ 
				$1/16$ & 75.1 &  91.7 & 30.6 \\ 
				$1/32$ & 74.2 & 91.3 & 28.6 \\ 
		\end{tabular}}\hspace{3mm}
		\subfloat[\textbf{Weaker spatial input to Fast pathway}: Alternative ways of weakening spatial inputs to the Fast pathway in SlowFast models. $\phi$$=$$1/8$ unless specified otherwise. 
		\label{tab:ablation:modality}]{
			\tablestyle{2pt}{1.05}
			\begin{tabular}{c|x{28}|x{22}x{22}x{22}}
				\multicolumn{1}{c|}{Fast pathway } & spatial & top-1 & top-5 & GFLOPs \\
				\shline
				RGB  & - & \textbf{75.6} & \textbf{92.1} & 36.1 \\ 
				\hline
				RGB,  $\phi$=$1/4$   & \emph{half}  & 74.7 &  91.8 & 34.4 \\ 
				gray-scale    & -  &  \textbf{75.5} & \textbf{91.9} & \textbf{34.1} \\ 
				time diff  & -  & 74.5 & 91.6 & 34.2 \\ 
				optical flow  & -  & 73.8 & 91.3 & 35.1 \\ 
				\multicolumn{5}{c}{} \\ 
				\multicolumn{5}{c}{} \\ 
			\end{tabular}
		}
		
		\vspace{0.5em}
		\caption{{Ablations} on  the Fast pathway design on \textbf{Kinetics-400}. We show top-1 and top-5 classification accuracy (\%), as well as computational complexity measured in GFLOPs (floating-point operations, in \# of multiply-adds \x $10^9$) for a  single clip input of spatial size 256$^2$. Inference-time computational cost is proportional to this, as a fixed number of 30 of views is used. Backbone: 4\x 16, R-50.}
		\label{tab:ablations}
		\vspace{-0.8em}
	\end{table*}
	
	\paragraph{Charades} \cite{Sigurdsson2016} is a dataset with longer range activities. \tblref{tab:sota:charades} shows our SlowFast results on it. For fair comparison, our baseline is the Slow-only counterpart that has 39.0 mAP. SlowFast increases over this baseline by 3.1 mAP (to 42.1), while the extra NL leads to an additional 0.4 mAP. We also achieve 45.2 mAP when pre-trained on Kinetics-600. Overall, our SlowFast models in \tblref{tab:sota:charades} outperform the previous best number (STRG \cite{Wang2018b}) by solid margins, at lower cost.
		
	\subsection{Ablation Experiments}
	This section provides ablation studies on Kinetics-400 comparing accuracy and computational complexity. 
	
	\paragraph{Slow \vs SlowFast.}
	We first aim to explore the SlowFast complementarity by changing the sample rate  ($T$\x$\tau$) of the Slow pathway. 
	Therefore, this ablation studies $\omega$, the frame rate ratio between the Fast and Slow paths. 	\figref{fig:slowVsSlowFast} shows the accuracy \vs complexity tradeoff for various instantiations of Slow and SlowFast models. It is seen that doubling the number of frames in the Slow pathway increases performance (vertical axis) at double computational cost (horizontal axis), while SlowFast significantly extends the performance of all variants at small increase of computational cost, even if the Slow pathways operates on higher frame rate. Green arrows illustrate the gain of adding the Fast pathway to the corresponding Slow-only architecture. The red arrow illustrates that SlowFast provides higher accuracy \textit{and} reduced cost. 
	
	Next, Table~\ref{tab:ablations} shows a series of ablations on the Fast pathway design, using the default SlowFast, $T$\x$\tau$ $=$ 4\x16, R-50 instantiation (specified in \tblref{tab:arch}), analyzed in turn.
	
	\paragraph{Individual pathways.} The first two rows in \tblref{tab:ablation:fusion} show the results for using the structure of one individual pathway \emph{alone}. The default instantiations of the Slow and Fast pathway are very lightweight with only 27.3 and 6.4 GFLOPs, 32.4M and 0.53M parameters, producing 72.6\% and 51.7\% top-1 accuracy, respectively. The pathways are designed with their special expertise if they are used jointly, as is ablated next.
	
	\paragraph{SlowFast fusion.} \tblref{tab:ablation:fusion} shows various ways of fusing the Slow and Fast pathways.
	As a na\"ive fusion baseline, we show a variant using no lateral connection: it only concatenates the final outputs of the two pathways. This variant has 73.5\% accuracy, slightly better than the Slow counterpart by 0.9\%.
	
	Next, we ablate SlowFast models with various lateral connections: time-to-channel (TtoC), time-strided sampling (T-sample), and time-strided convolution (T-conv). For TtoC, which can match channel dimensions, we also report fusing by element-wise summation (TtoC, sum). For all other variants concatenation is employed for fusion.
	
	\tblref{tab:ablation:fusion} shows that these SlowFast models are \emph{all} better than the Slow-only pathway. With the best-performing lateral connection of T-conv, the SlowFast network is \textbf{3.0\% better} than Slow-only. We employ T-conv as our default.
	
	Interestingly, the Fast pathway alone has only 51.7\% accuracy (\tblref{tab:ablation:fusion}). But it brings in up to 3.0\% improvement to the Slow pathway, showing that the underlying representation modeled by the Fast pathway is largely complementary. We strengthen this observation by the next set of ablations.
	
	\paragraph{Channel capacity of Fast pathway.} A key intuition for designing the Fast pathway is that it can employ a lower channel capacity for capturing motion \emph{without} building a detailed spatial representation. This is controlled by the channel ratio  $\phi$. \tblref{tab:ablation:widthFast} shows the effect of varying $\phi$.
	
	The best-performing $\phi$ values are $1/6$ and $1/8$ (our default). Nevertheless, it is surprising to see that \emph{all} values from $\phi$$=$$1/32$ to $1/4$ in our SlowFast model can improve over the Slow-only counterpart. In particular, with $\phi$$=$$1/32$, the Fast pathway only adds as small as 1.3 GFLOPs ($\app$5\% relative), but leads to 1.6\% improvement.

\begin{table}[t!]\centering  
	\tablestyle{1.7pt}{1.05}
	\begin{tabular}{l|c|x{22}x{28}x{22}x{22}x{28}}
		\multicolumn{1}{c|}{model} & pre-train & top-1 & top-5 & GFLOPs \\
		\shline
		\demph{3D R-50  \cite{Wang2018}} & \demph{ImageNet}    & \demph{73.4} & \demph{90.9} & \demph{36.7} \\ 
		3D R-50, recipe in \cite{Wang2018} & -    & 69.4 & 88.6 & 36.7\\
		3D R-50, our recipe & -     & 73.5 & 90.8 & 36.7\\
\end{tabular}

\caption{\textbf{Baselines trained from scratch}: 
	Using the same network structure as \cite{Wang2018}, our training recipe achieves comparable results \emph{without} ImageNet pre-training.}
\label{tab:ablations:imagenet}
\vspace{-0.8em}
\end{table}

	\paragraph{Weaker spatial inputs to Fast pathway.} Further, we experiment with using different \emph{weaker} spatial inputs to the Fast pathway in our SlowFast model. We consider: (i) a \emph{half spatial resolution} (112\x112), with $\phi$$=$$1/4$ (\vs default $1/8$) to roughly maintain the FLOPs; (ii) \emph{gray-scale} input frames; (iii) ``\emph{time difference}" frames, computed by subtracting the current frame with the previous frame; and (iv) using \emph{optical flow} as the input to the Fast pathway.
	
	\tblref{tab:ablation:modality} shows that all these variants are competitive and are better than the Slow-only baseline. In particular, the \emph{gray-scale} version of the Fast pathway is nearly as good as the RGB variant, but reduces FLOPs by $\app$5\%. Interestingly, this is also consistent with the M-cell's behavior of being insensitive to colors \cite{Hubel1965,Livingstone1988,Derrington1984,Felleman1991,VanEssen1994}.
	
	We believe both \tblref{tab:ablation:widthFast} and \tblref{tab:ablation:modality} convincingly show that the \emph{lightweight} but temporally \emph{high-resolution} Fast pathway is an effective component for video recognition.

	\paragraph{Training from scratch.} Our models are trained \emph{from scratch}, without ImageNet training. To draw fair comparisons, it is helpful to check the potential impacts (positive or negative) of training from scratch. To this end, we train \emph{the exact same} 3D ResNet-50 architectures specified in \cite{Wang2018}, using our large-scale SGD recipe trained from scratch.
	
	\tblref{tab:ablations:imagenet} shows the comparisons using this 3D R-50 baseline architecture. We observe, that our training recipe achieves \emph{comparably good} results as the ImageNet pre-training counterpart reported by \cite{Wang2018}, while the recipe in \cite{Wang2018} is not well tuned for directly training from scratch. This suggests that our training system, as the foundation of our experiments, has no loss for this baseline model, despite not using ImageNet for pre-training.

	\section{Experiments: AVA Action Detection}\label{sec:detection}
	
\begin{figure*}[t!]	\vspace{-1em}
		\centering
		\vspace{-1.8em}
		\includegraphics[width=1\textwidth]{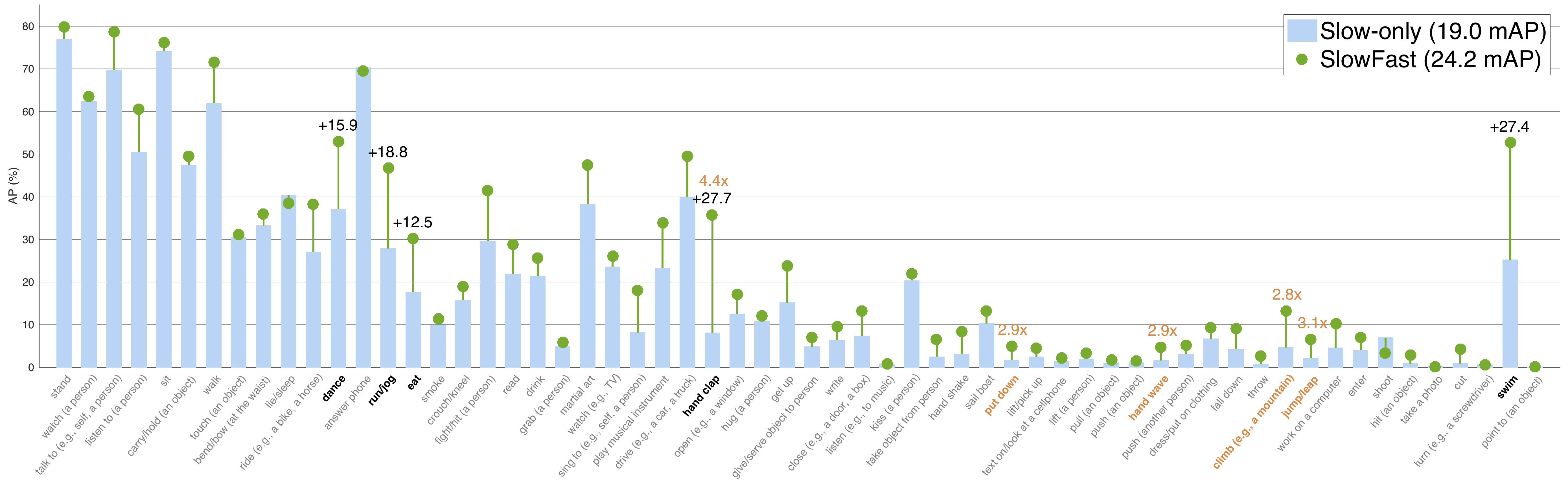}
		\vspace{-1.0em}
		\caption{\textbf{Per-category AP on AVA}: a Slow-only baseline (19.0 mAP) \vs its SlowFast counterpart (24.2 mAP). The highlighted categories are the 5 highest absolute increase (\textbf{black}) or 5 highest relative increase with Slow-only AP $>$ 1.0 ({\color{orange}{\textbf{orange}}}). Categories are sorted by number of examples. Note that the SlowFast instantiation in this ablation is not our best-performing model. }
		\vspace{-.5em}
		\label{fig:ava:class_ap}
	\end{figure*}

	\paragraph{Dataset.}
	The AVA dataset \cite{Gu2018} focuses on spatiotemporal localization of human actions.
	The data is taken from 437 movies. Spatiotemporal labels are provided for one frame per second, with every person annotated with a bounding box and (possibly multiple) actions. Note the difficulty in AVA lies in action detection, while actor localization is less challenging \cite{Gu2018}. 
	There are 211k training and 57k validation video segments in AVA v2.1 which we use. We follow the standard protocol  \cite{Gu2018} of evaluating on 60 classes (see \figref{fig:ava:class_ap}).
	The performance metric is mean Average Precision (mAP) over 60 classes, using a frame-level IoU threshold of 0.5. 
	
	\paragraph{Detection architecture.}
		Our detector is similar to Faster R-CNN \cite{Ren2015} with minimal modifications adapted for video.
	We use the SlowFast network or its variants as the backbone.
	We set the spatial stride of res$_5$ to 1 (instead of 2), and use a dilation of 2 for its filters. This increases the spatial resolution of res$_5$ by 2$\times$.
	We extract region-of-interest (RoI) features \cite{Girshick2015} at the last feature map of res$_5$.
	We first extend each 2D RoI at a frame into a 3D RoI by replicating it along the temporal axis, similar to the method presented in \cite{Gu2018}. Subsequently, we compute RoI features by RoIAlign \cite{He2017} spatially, and global average pooling temporally. The RoI features are then max-pooled and fed to a per-class, sigmoid-based classifier for multi-label prediction.
	
	We follow previous works that use pre-computed proposals \cite{Gu2018,Sun2018,Jiang2018}. Our region proposals are computed by an off-the-shelf person detector, \ie, that is not jointly trained with the action detection models.
	We adopt a person-detection model trained with \emph{Detectron} \cite{Detectron2018}. It is a Faster R-CNN with a ResNeXt-101-FPN~\cite{Xie2017,Lin2017} backbone.
	It is pre-trained on ImageNet and the COCO human keypoint images~\cite{Lin2014}.
	We fine-tune this detector on AVA for person (actor) detection.
	The person detector produces 93.9 AP@50 on the AVA validation set.
	Then, the region proposals for action detection are detected person boxes with a confidence of $>$ 0.8, which has a recall of 91.1\% and a precision of 90.7\% for the person class.
	
	\paragraph{Training.} We initialize the network weights from the Kinetics-400 classification models.
	We use step-wise learning rate, reducing the learning rate 10\x~when validation error saturates. 
	We train for 14k iterations (68 epochs for $\app$211k data), with linear warm-up \cite{Goyal2017} for the first 1k iterations. We use a weight decay of 10$^{-7}$. 
	All other hyper-parameters are the same as in the Kinetics experiments.
	Ground-truth boxes are used as the samples for training.
		The input is instantiation-specific $\alpha T$\x$\tau$ frames of size 224\x224.
	
	\paragraph{Inference.} We perform inference on a single clip with $\alpha T$\x$\tau$ frames around the frame that is to be evaluated. We resize the spatial dimension such that its shorter side is 256 pixels. The backbone feature extractor is computed fully convolutionally, as in standard Faster R-CNN \cite{Ren2015}.
	
	\begin{table}[t!]
		\vspace{-0.8em}
		\centering
		\tablestyle{2.5pt}{1.05}
		\begin{tabular}{l|c|c|c|c}
			\multicolumn{1}{c|}{model} & flow &\multicolumn{1}{c|}{video pretrain} &   val mAP  & test mAP  \\ 
			\shline
			I3D   \cite{Gu2018} & & Kinetics-400 & 14.5  & -\\
			I3D  \cite{Gu2018} & \checkmark& Kinetics-400 & 15.6  & -\\
			ACRN,  S3D \cite{Sun2018} &\checkmark& Kinetics-400 & 17.4  & -\\
			ATR, R50+NL \cite{Jiang2018}  & &Kinetics-400 &  20.0  & -\\
			ATR, R50+NL  \cite{Jiang2018}    &\checkmark& Kinetics-400 & 21.7  & -\\
			\demph{9-model ensemble \cite{Jiang2018}}  &\checkmark & \demph{Kinetics-400} & \demph{25.6} & \demph{21.1}\\
			I3D \cite{Girdhar2018}\quad\ &  &  Kinetics-600 &  21.9 & 21.0  \\
			\hline
			\textbf{SlowFast} & & Kinetics-400   & 26.3  & - \\
			\textbf{SlowFast} & & Kinetics-600  & {26.8}  & - \\ 
			\textbf{SlowFast}, +NL & & Kinetics-600  & 27.3  & \textbf{27.1} \\ 
			\textbf{SlowFast}*, +NL && Kinetics-600  & \textbf{28.2}  & - \\ 
		\end{tabular}
		\vspace{.5em}
		\caption{\textbf{Comparison with the state-of-the-art on AVA v2.1}.  All our variants are based on  $T$\x$\tau$  $=$ 8\x8, R101. 
			Here ``*'' indicates a version of our method that uses our region proposals for training.
		}
		\label{tab:ava:sota}
	\end{table}

	\subsection{Main Results}
	
	We compare with previous results on AVA in \tblref{tab:ava:sota}. An interesting observation is on the potential benefit of using optical flow (see column `flow' in \tblref{tab:ava:sota}). Existing works have observed mild improvements: +1.1 mAP for I3D in \cite{Gu2018}, and +1.7 mAP for ATR in \cite{Jiang2018}. In contrast, our baseline improves by the Fast pathway by +5.2 mAP (see \tblref{tab:ava:base} in our ablation experiments in the next section). Moreover, two-stream methods using optical flow can \emph{double} the computational cost, whereas our Fast pathway is lightweight.
	
	As system-level comparisons, our SlowFast model has 26.3 mAP using only Kinetics-400 pre-training. This is \textbf{5.6} mAP higher than the previous best number under similar settings (21.7 of ATR \cite{Jiang2018}, single-model), and \textbf{7.3} mAP higher than that using no optical flow (\tblref{tab:ava:sota}). 
	
	The work in \cite{Girdhar2018} pre-trains on the larger Kinetics-600 and achieves 21.9 mAP. For fair comparison, we observe an improvement from 26.3 mAP to 26.8 mAP for using Kinetics-600. Augmenting SlowFast with NL blocks \cite{Wang2018} increases this to 27.3 mAP. 
	We train this model on train+val (and by 1.5$\times$ longer) and submit it to the AVA v2.1 test server \cite{AVALeaverboard2018}. It achieves \textbf{27.1~mAP} single crop test set accuracy. 
	
	By using predicted proposals overlapping with ground-truth boxes by IoU $>$ 0.9, in addition to the ground truth boxes, for training we achieve \textbf{28.2 mAP} single crop validation accuracy, a new state-of-the-art on AVA. 
	
	\begin{table}[t!]\vspace{-.8em}
		\centering
		\tablestyle{2.5pt}{1.05}
		\begin{tabular}{l|c|c|c|c}
			\multicolumn{1}{c|}{model} & flow &\multicolumn{1}{c|}{video pretrain} &   val mAP  & test mAP  \\ 
			\shline
			\textbf{SlowFast}, 8\x8 &  &Kinetics-{600}  & {29.0} & - \\
			\textbf{SlowFast}, 16\x8 & &Kinetics-{600}  & {29.8} & - \\
			\textbf{SlowFast}++, 16\x8 &  &Kinetics-{600}  & \textbf{30.7} & - \\
			\textbf{SlowFast}++, ensemble &  &  Kinetics-{600}  & - & \textbf{34.3}  \\
		\end{tabular}
		\vspace{.5em}
		\caption{\textbf{SlowFast models on AVA v2.2}.
			Here ``++'' indicates a version of our method that is tested with multi-scale and horizontal flipping augmentation. The backbone is R-101+NL and region proposals are used for training.
		}
		\label{tab:ava:challenge}
		\vspace{-1.2em}
	\end{table}

	Using the AVA v2.2 dataset (which provides more consistent annotations) improves this number to 29.0 mAP (\tblref{tab:ava:challenge}). The longer-term \textbf{SlowFast}, 16\x8 model produces 29.8 mAP and using multiple spatial scales and horizontal flip for testing, this number is increased to \textbf{30.7 mAP}.
	
	Finally, we create an ensemble of 7 models and submit it to the official test server for the ActivityNet challenge 2019 \cite{ActivityNet2019}. As shown in \tblref{tab:ava:challenge} this entry (\textbf{SlowFast}++, ensemble) achieved \textbf{34.3 mAP} accuracy on the test set, ranking first in the AVA action detection challenge 2019. Further details on our winning solution are provided in the corresponding technical report \cite{Feichtenhofer2019a}.

	\subsection{Ablation Experiments}\label{sec:ava:ablations}
	
	\tblref{tab:ava:base} compares a Slow-only baseline with its SlowFast counterpart, with the \emph{per-category} AP shown in \figref{fig:ava:class_ap}. Our method improves massively by \textbf{5.2} mAP (relative 28\%) from 19.0 to 24.2. This is \emph{solely} contributed by our SlowFast idea.

	Category-wise (\figref{fig:ava:class_ap}), our SlowFast model improves in \textbf{57 out of 60} categories, \vs its Slow-only counterpart.
	The largest absolute gains are observed for ``\emph{hand clap}" (+27.7 AP), ``\emph{swim}" (+27.4 AP), ``\emph{run/jog}" (+18.8 AP), ``\emph{dance}" (+15.9 AP), and ``\emph{eat}'' (+12.5 AP).
	We also observe large relative increase in ``\emph{jump/leap}'', ``\emph{hand wave}'', ``\emph{put down}'', ``\emph{throw}'', ``\emph{hit}'' or ``\emph{cut}''.
	These are categories where modeling dynamics are of vital importance.
	The SlowFast model is worse in only 3 categories: ``\emph{answer phone}" (-0.1 AP), ``\emph{lie/sleep}" (-0.2 AP), ``\emph{shoot}" (-0.4 AP), and their decrease is relatively small \vs others' increase.

	\section{Conclusion}
	
	The time axis is a special dimension.
	This paper has investigated an architecture design that contrasts the speed along this axis. It achieves state-of-the-art accuracy for video action classification and detection. 
	We hope that this SlowFast concept will foster further research in video recognition.

\begin{table}[t] 
	\centering
	\tablestyle{3pt}{1.05} 
	\begin{tabular}{l|x{32}x{10}|x{22}}
		\multicolumn{1}{c|}{ model } & ${T \times \tau}$ & $\omega$   & mAP \\
		\shline
		Slow-only, R-50   &4\x16  & - & 19.0 \\ 
		SlowFast, R-50  & 4\x16 & 8  & \textbf{24.2}  \\  
	\end{tabular}
	\vspace{.3em}
	\caption{\textbf{AVA action detection baselines}: Slow-only \vs SlowFast.}
	\label{tab:ava:base}
\end{table}

	\appendix
	\section{Appendix} \label{sec:appendix}
\vspace{-.3em}
\paragraph{Implementation details.}
	We study backbones including ResNet-50 and the deeper ResNet-101 \cite{He2016}, optionally augmented with non-local (NL) blocks \cite{Wang2018}. For models involving R-101, we use a scale jittering range of [256, 340]. The $T$\x$\tau$  $=$ 16\x8 models are initilaized from the 8\x8 counterparts and trained for half the training epochs to reduce training time. For all models involving NL, we initialize them with the counterparts that are trained without NL, to facilitate convergence. We only use NL on the (fused) Slow features of res$_4$ (instead of res$_3$+res$_4$ \cite{Wang2018}). 

On \emph{Kinetics}, we adopt synchronized SGD training in 128 GPUs following the recipe in \cite{Goyal2017}, and we found its accuracy is as good as typical training in one 8-GPU machine but it scales out well. 
	The mini-batch size is 8 clips per GPU (so the total mini-batch size is 1024).
We use the initialization method in \cite{He2015}.
	We train with Batch Normalization (BN) \cite{Ioffe2015} with BN statistics computed within each 8 clips.
	We adopt a half-period cosine schedule \cite{Loshchilov2016} of learning rate decaying: the learning rate at the $n$-th iteration is $\eta\cdot0.5[\cos(\frac{n}{n_\text{max}}\pi)+1]$, where $n_\text{max}$ is the maximum training iterations and the base learning rate $\eta$ is set as 1.6.
	We also use a linear warm-up strategy \cite{Goyal2017} in the first 8k iterations.
For Kinetic-400, we train for 256 epochs (60k iterations with a total mini-batch size of 1024, in $\app$240k Kinetics videos) when $T\leq$ 4 frames, and 196 epochs when $T>$ 4 frames: it is sufficient to train shorter when a clip has more frames.
	We use momentum of 0.9 and weight decay of 10$^\text{-4}$. Dropout \cite{Hinton2012b} of 0.5 is used before the final classifier layer.
 
 	For \emph{Kinetics-600}, we extend the training epochs (and schedule) by 2\x~and set the base learning rate $\eta$ to 0.8.

 	For \emph{Charades}, we fine-tune the Kinetics models. A per-class sigmoid output is used to account for the mutli-class nature. We train on a single machine for 24k iterations using a batch size of 16 and a base learning rate of 0.0375 (Kinetics-400 pre-trained) and 0.02 (Kinetics-600  pre-trained) with 10\x~step-wise decay if the validation error saturates. For inference, we temporally max-pool scores \cite{Wang2018}. 

	{
		\small
		\bibliographystyle{ieee}
		\bibliography{slowfast}
	}

\end{document}